\documentclass[10pt,letterpaper]{article}
\usepackage[top=0.85in,left=2.75in,footskip=0.75in,marginparwidth=2in]{geometry}

\usepackage[utf8]{inputenc}

\usepackage{cite}

\usepackage{algorithm2e}

\usepackage{nameref,hyperref}

\usepackage[right]{lineno}

\usepackage{microtype}
\DisableLigatures[f]{encoding = *, family = * }

\raggedright
\usepackage{changepage}

\usepackage[aboveskip=1pt,labelfont=bf,labelsep=period,singlelinecheck=off]{caption}

\makeatletter
\renewcommand{\@biblabel}[1]{\quad#1.}
\makeatother

\usepackage{lastpage,fancyhdr,graphicx}
\usepackage{epstopdf}
\fancyheadoffset[L]{2.25in}
\fancyfootoffset[L]{2.25in}

\usepackage{color}

\definecolor{Gray}{gray}{.25}

\usepackage{graphicx}
\usepackage{multirow}
\usepackage{amsmath}
\usepackage{amssymb}
\usepackage{sidecap}

\usepackage{wrapfig}
\usepackage[pscoord]{eso-pic}
\usepackage[fulladjust]{marginnote}
\reversemarginpar

\begin{document}
\vspace*{0.35in}

% sharp symbol
%newcommand{CS}{Cnolinebreak[4]hspace{-0.05em}raisebox{.4ex}{relsize{-2}{textbf{#}}}}

% title goes here:
\begin{flushleft}
{\Large
\textbf\newline{Optimization of Convolutional Neural Network using Microcanonical Annealing Algorithm}
}
\newline
% authors go here:
\\
Vina Ayumi\textsuperscript{1},
L.M. Rasdi Rere\textsuperscript{1,2},
Mohamad Ivan Fanany\textsuperscript{1},
Aniati Murni Arymurthy\textsuperscript{1},
%Author 4\textsuperscript{1},
%Author 5\textsuperscript{2},
%Author 6\textsuperscript{2},
%Author 7\textsuperscript{1,*}
\\
\bigskip
\bf{1} Machine Learning and Computer Vision Laboratory, \\Faculty of Computer Science, Universitas Indonesia\\
\bf{2} Computer System Laboratory, STMIK Jakarta STI\&K\\
\bigskip
* vina.ayumi@ui.ac.id

\end{flushleft}

\providecommand{\keywords}[1]{\textbf{\textit{Keywords---}} #1}

\section*{Abstract}
Convolutional neural network (CNN) is one of the most prominent architectures and algorithm in Deep Learning. It shows a remarkable improvement in the recognition and classification of objects. This method has also been proven to be very effective in a variety of computer vision and machine learning problems. As in other deep learning, however, training the CNN is interesting yet challenging. Recently, some metaheuristic algorithms have been used to optimize CNN using Genetic Algorithm, Particle Swarm Optimization, Simulated Annealing and Harmony Search. In this paper, another type of metaheuristic algorithms with different strategy has been proposed, i.e. Microcanonical Annealing to optimize Convolutional Neural Network. The performance of the proposed method is tested using the MNIST and CIFAR-10 datasets. Although experiment results of MNIST dataset indicate the increase in computation time (1.02x - 1.38x), nevertheless this proposed method can considerably enhance the performance of the original CNN (up to 4.60\%). On the CIFAR10 dataset, currently, state of the art is 96.53\% using fractional pooling, while this proposed method achieves 99.14\%. 
\bigskip

\noindent\keywords{Metaheuristic, Microcanonical Annealing, Convolutional Neural Network, MNIST, CIFAR10}

% now start line numbers
%\linenumbers

\section{Introduction}
\label{sec1}
Essentially, Deep learning (DL) is motivated by the artificial intelligent (AI) research, where the objective is to replicate the human brain capability, i.e. to observe, learn, analyze and make a decision, particularly for complex problems \cite{Naja}. DL is about learning the representation of a hierarchical feature, and it contains a variety of methods, such as neural network, hierarchical of probabilistic models, and supervised as well as unsupervised learning algorithms.\cite{Liangpei}. The current good reputation of DL is due to the decrease in the price of computer hardware, improvement in the computational processing capabilities, and advanced research in the Machine Learning and Signal Processing \cite{Deng}.

In general, DL models can be classified into discriminative, generative, and hybrid models\cite{Deng}. Recurrent neural network (RNN), deep neural networks (DNN), and convolutional neural networks (CNN) are some examples of Discriminative models. Examples of generative models are deep Boltzmann machine (DBM), regularized autoencoder, and deep belief network (DBN). In the case of the hybrid model, it refers to a combination of generative and discriminative models. An example of such hybrid model is a pre-trained deep CNN using DBN, where it can improve the performance of deep CNN better than if it uses only random initialization. Among all of these DL techniques, this paper focuses on CNN. 

Although the good reputation of DL for solving any learning problem is known, how to train it is challenging. The successful proposal to optimize this technique using layered-wise pre-training was proposed by Hinton and Salakhutdinov \cite{Hinton}. Some other methods are Hessian-free optimization suggested by Marten \cite{Martens}, and Krylov Subspace Descent by Vinyal et al. \cite{Vinyal}

Recently, some of the metaheuristic algorithms have been used to optimize DL, especially CNN. Some papers\cite{You}\cite{Oul}\cite{Rasdi}\cite{Gustavo} \cite{Laode} report that these methods can improve the accuracy of CNN. Metaheuristic is a powerful method to solve difficult optimization problems, and it has been used in almost all research area of engineering, science, and even industrial application \cite{Yang}. In general, this method works with three main objectives, i.e. solving big problems, solving the problem faster, and finding robust algorithms \cite{Talbi}. Besides, they are not difficult to be designed, flexible, and relatively easy to be applied.

Almost all metaheuristics algorithms inspired by nature, which is based on several principles of phenomena in physics, biology, and ethology. Some examples of biology phenomena are Differential Evolution (DE), Evolution Strategy (ES), Genetic Algorithm (GA). Phenomena of physics are Threshold Accepting method (TA), Microcanonical Annealing (MA), Simulated Annealing (SA), and Ethology phenomena are  Ant Colony Optimization (ACO), Firefly Algorithm (FA), Particle Swarm Optimization (PSO)\cite{Boussaid}. Another metaheuristic phenomenon is inspired by music, such as Harmony Search algorithm \cite{Lee}. 

Classifications of metaheuristic can also be based on single-solution based metaheuristic (S-metaheuristic) and population-based metaheuristic (P-metaheuristic). Examples of S-metaheuristic are SA, TA, MA, Guided Local Search, and Tabu Search. In the case of P-metaheuristic, it can be divided into Swarm Intelligent (SI) and Evolutionary Computation (EC). Examples of SI are FA, PSO, ACO, Bee Colony Optimization and examples of EC are GA, ES, DE \cite {Boussaid}. 

Of the various types of the metaheuristic algorithm, in this paper we use the MA, with the consideration that the S-Metaheuristic is simple to implement on DL, and to the best of our knowledge, has never been used for optimizing CNN. 

The Macrocanonic algorithm is the variant of Simulated Annealing. Uses an adaptation of the  Metropolis algorithm, the conventional SA algorithm aims to bring a system to equilibrium at decreasing temperatures \cite{Stephen}. On the other hand, MA based on Creutz’s microcanonical simulation technique, where the system’s evolution is controlled by its internal energy, not by its temperature. The advantages Creutz algorithm over the Metropolis algorithm is since it does not require the generation of quality random numbers or the evaluation of transcendental functions, thus allowing much faster implementation. Experiments on the Creutz method indicate that, it can be programmed to run an order of magnitude faster than the conventional Metropolis method for discrete systems \cite{Bhanot}. A further significant advantage is that microcanonical simulation does not require high-quality random numbers.

The organization of this paper is as follows: Section 1 is an introduction; Section 2 provides an overview of Microcanonical Annealing; Section 3 describes the method convolutional neural network; Section 4 presents the proposed method; Section 5 gives the results of the experiment, and lastly, Section 6 presents the conclusion of this paper.

%+++++++++++++++++++++++++++++++++++++++++++++++++++++++
\section{Microcanonical Annealing}
%+++++++++++++++++++++++++++++++++++++++++++++++++++++++

Microcanonical Annealing (MA) corresponds to a variant of simulated annealing (SA). This technique is based on the Creutz algorithm, known as “demon” algorithm or microcanonical Monte Carlo simulation. In which the algorithm tolerates attainment of the equilibrium of thermodynamic in an isolated system, where in this condition, total energy of the system \(E_p\) is constant \cite{Boussaid}. 

Total energy is the sum of kinetic energy \(E_k\) and potential energy \(E_p\) of the system, as the equation (2) follow:

\begin{equation}
\label{eq1}
   E_{total} = E_k + E_p  
\end{equation}

In case of minimum optimization problem, potential energy \(E_p\) is the objective function to be minimized, and the kinetic energy is used as temperature in SA, that is forced to remain positive \cite{Boussaid}. When the change of energy is negative
\((- \Delta E)\), while it increases the kinetic energy \((E_k \leftarrow E_k - \Delta E  )\), this new states is accepted. Otherwise, it is accepted when \(- \Delta E < E_k\), and the energy obtained in the form of potential energy is cut off from the kinetic energy. So that the total energy remains constant. The standard algorithm for MA is shown in \textbf{Algorithm 1}.\cite{Boussaid}.

\begin{algorithm}[H]
\SetAlgoLined
\KwResult{accuracy \(A_1\), computational time \(T_1\)}
 Randomly, select an initial solution $x$\;
 Initialize the kinetic energy \(E_k\)\; 
 
 \While{termination criteria is not satisfied}{
 \Repeat{thermodynamic equilibrium achieved on the system}{
  Randomly select \( x' \in N(x)\)\;
  Calculate \( \Delta E = E(x') - E(x)\) \;
  
  \If{\( E(x') < E(x)\)}{
   \( x  \leftarrow  x'\)\;
   \( E_k \leftarrow  E_k + \Delta x\)\;
  }
  }
  decrease the temperature: \(T=c \times T\) \;
  update $x$  for all layer;
 }
 
 \caption{Microcanonical Annealing}
\end{algorithm}

%+++++++++++++++++++++++++++++++++++
\section{Convolutional neural network}
%+++++++++++++++++++++++++++++++++++
One variant of the standard multilayer perceptron (MLP) is CNN. Its capability in reducing the dimension of data, extracting the feature sequentially, and classifying in one structure of network are distinguished advantages of this method, especially, for pattern recognition compared with the conventional approaches  \cite{Bengio}. 

The classical CNN by LeCun et al \cite{LeCun} is an extension of traditional MLP based on three ideas: local receive fields, weights sharing, and spatial/temporal sub-sampling. There are two types of processing layers, which are convolution layers and sub-sampling layers. As demonstrated in Fig.1, the processing layers contain three convolution layers C1, C3, and C5, combined in between with two sub-sampling layers S2 and S4, and output layer F6. These convolution and sub-sampling layers are arranged into planes called features maps.

In convolution layer, each neuron is locally linked to a small input region (local receptive field) in the preceding layer. All neurons with similar feature maps obtain data from different input regions until the whole plane input is skimmed, but the similar weights are used together (weights sharing).

\begin{figure}
%\centering
\includegraphics[scale = 0.55]{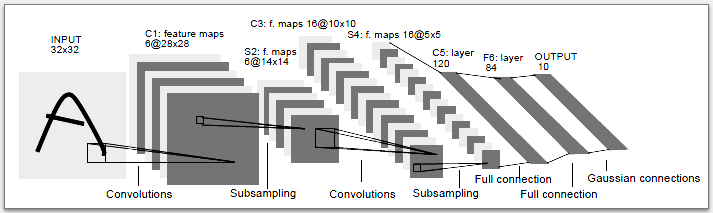}
\caption{Architecture of CNN by LeCun et al (LeNet5)}
\label{fig:my_label}
\end{figure}

The feature maps are spatially down-sampled in sub-sampling layer, in which the map size is reduced by a factor 2. For instance, the feature map in layer C3 of size 10x10 is sub-sampled to a conforming feature map of size 5x5 in the subsequent layer S4. The last layer is F6 that is the process of classification \cite{LeCun}.

Basically, a convolution layer is correlated with some feature maps, the size of the kernel, and connections to the previous layer. Each feature map is the result of a sum of convolution from the maps of the previous layer, by their corresponding kernel and a linear filter. Furthermore, a bias term is added to the map then and applying it to a non-linear function. The k-th feature map \(M_{ij}^k\) with the weights \(W^k\) and bias \(b_k\) is obtained using the $\tanh$ function as follow:

\begin{equation}
M_{ij}^k=\tanh((W^k \times x)_{ij} + b_k)
\end{equation}

The purpose of a sub-sampling layer is the spatially invariant reached by reducing the feature maps resolution, where each feature map is pooled relating to one of the feature map of the previous layer

where each map feature is collected relating to one of the maps of the features of the previous layer. Where \(a_i^{n \times n} \) are the inputs, \( \beta \) is a scalar of trainable, and \( b \) is bias of trainable, the sub-sampling function,is given by the following equation:

\begin{equation}
a_j=\tanh\left(\beta\sum_{N\times N}{a_i^{n \times n} + b}\right)
\end{equation}

After several convolutions and sub-samplings, the last structure is a classification layer. This layer works as an input for a series of fully connected layers that will execute the classification task. In this layer, each output neuron is assigned to one class label, and in the case of CIFAR10 or MNIST data set, this layer contains ten neurons corresponding to their classes.

%+++++++++++++++++++++++++++++++++++ 
\section{Design of proposed methods}
%+++++++++++++++++++++++++++++++++++

In this proposed method, the algorithm of MA is used to train CNN to find the condition of best accuracy, as well as to minimize estimated error and indicator of network complexity. This objective can be realized by computing the loss function of vector solution or the standard error on the training set. The following is the loss function used in this paper:
\begin{equation}
f= \frac {1}{2} \left({\frac{\sum_{i=N}^{N}{(x - y)^2}}{N}}\right)^{0.5}
\end{equation}
where the expected output is $x$, the real output is $y$, and some of the training samples are $N$. The two situations are used in this method for termination criterion. The first is when the maximum iteration has been reached and the second is when the loss function is less than a certain constant. Both conditions mean that the most optimal state has been achieved.

The architecture of this proposed method is i-6c-2s-12c-2s, where the number of C1 is 6, and C3 is 12. The size of kernel for all convolution layer is 5x5,  and the scale of sub-sampling is 2. This architecture is a simple CNN structure (LeNet-5), not a complex structure like AlexNet\cite{Alexnet}, SPP\cite{KHe}, and GoogLeNet\cite{Szegedy}. In this paper, these architecture is designed for MNIST dataset.

Technically in these proposed methods, CNN will compute the values of bias and weight. These values ($x$) are used to calculate the loss function \(f(x)\). 

The values of $x$ are used as a vector of solution in MA, which will be optimized, by adding a value of \(\Delta x\) randomly. Meanwhile, \(f(x)\) is used as a potential energy $E_k$ in MA.

In this proposed method, \(\Delta x\) is one of the important parameters. The value of accuracy will be improved significantly by providing an appropriate value of the \(\Delta x\) parameter. As an example of one epoch, if \( \Delta x = 0.001 \times rand\), then the maximum accuracy is 87.60\%, in which this value is 5.21\% greater than the original CNN (82.39\%). However, if \( \Delta x = 0.0001 \times rand\), its accuracy is 85.45\% and its is only 3.06\% greater than the original CNN. 

Another important parameter of the proposed method is the size of neighborhood. For example in one epoch, if neighborhood is 5, 10 or 20, and then the accuracy values are respectively 85.74\%, 87.52\%, or 88.06\%. While the computing time are respectively 98.06 seconds, 99.18 seconds and 111.80 seconds.

Furthermore, this solution vector is updated based on MA algorithm. In case of termination criterion has been reached, all of biases and weights for all layers on the system will be updated. 

%+++++++++++++++++++++++++++
\section{Experiment and results}
%+++++++++++++++++++++++++++

In this paper, there are two categories of experiments conducted, based on the dataset. The first experiment was using MNIST dataset, and the  second experiment using CIFAR10 dataset. Some of the examples image for MNIST dataset are shown in Figure 2 and for CIFAR10 dataset are shown in Figure 3.

\begin{figure}
%\centering
\includegraphics[scale=0.6]{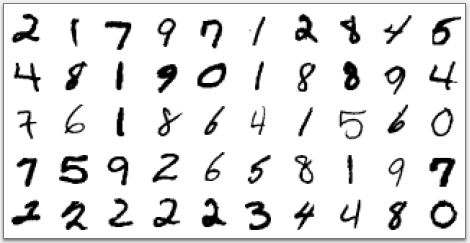}
\caption{Examples of some image from MNIST data-set}
\label{fig:my_label}
\end{figure}

\begin{figure}
%\centering
\includegraphics[scale = 0.7]{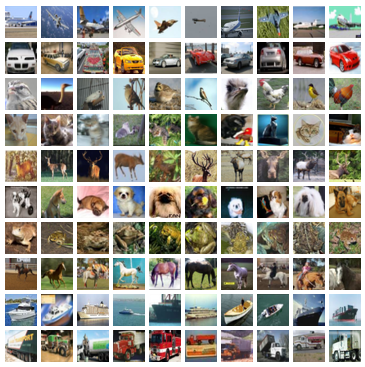}
\caption{Examples of some image from CIFAR10 data-set}
\label{fig:my_label}
\end{figure}

%-------------------------------------------
\subsection{Experiment using MNIST data set}
%-------------------------------------------

The experiment for MNIST data set was implemented in MATLAB-R2011a,  windows 10, on a PC with processor Intel Core i7-4500u, and 8 GB RAM running memory, with five experiments for each epoch. The original program of this experiment is DeepLearn Toolbox from Palm\cite{Palm}. In this research, the program of CNN is modified with the algorithm of MA.

In all experiment, the size of neighborhood was set to 10, maximum of iteration (maxit) = 10, as well as kinetic energy = 100. We also set the parameter of CNN i.e., the learning rate ($\alpha = 1$) and the batch size (100). 

On the MNIST dataset, all of the experiment results of the proposed methods are compared with the experiment result from the original CNN. The results of CNN and CNN based on MA is summarized in Table 1, for accuracy (A1, A2) and computation time (T1, T2), as well as Figure 4 for Error and Figure 5 for computation time. 

\begin{figure}
%\centering
\includegraphics[scale = 0.6]{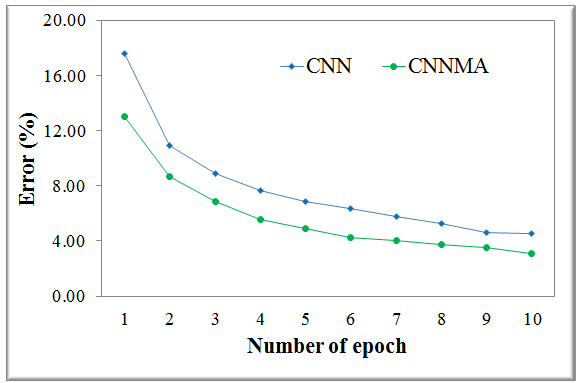}
\caption{Error for CNN and CNNMA}
\label{fig:my_label}
\end{figure}

\begin{figure}
%\centering
\includegraphics[scale = 0.6]{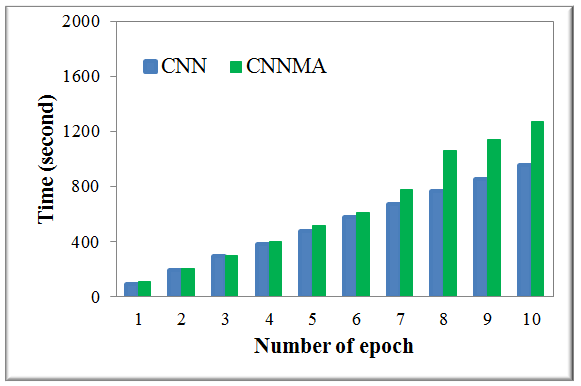}
\caption{Computation time for CNN and CNNMA}
\label{fig:my_label}
\end{figure}

In case of 100 epochs, as is shown in Figure 6 and 7, the accuracy of original CNN is 98.65\% and the accuracy of CNN by MA is 98.75\%. The computation time of both methods are 10731 seconds and 17090s seconds respectively.

\begin{figure}
%\centering
\includegraphics[scale = 0.4]{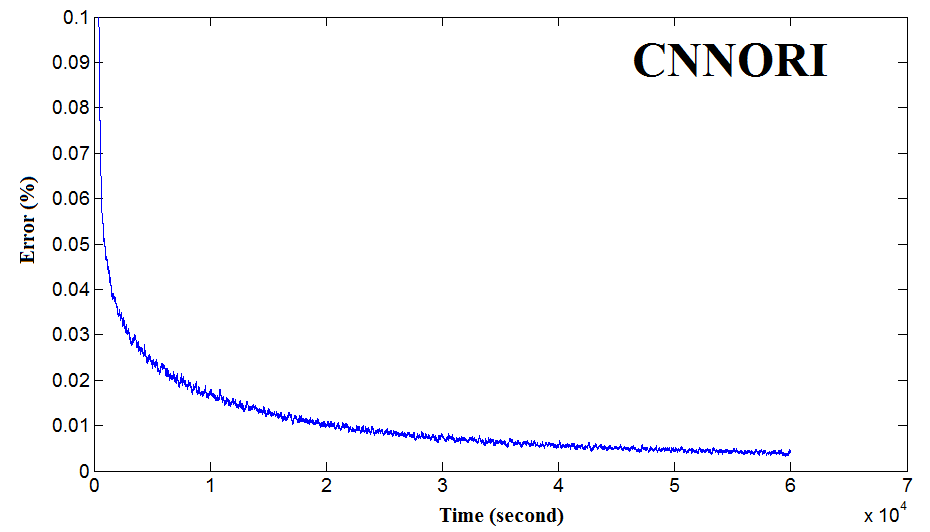}
\caption{Error vs Computation time for CNN}
\label{fig:my_label}
\end{figure}

\begin{figure}
%\centering
\includegraphics[scale = 0.44]{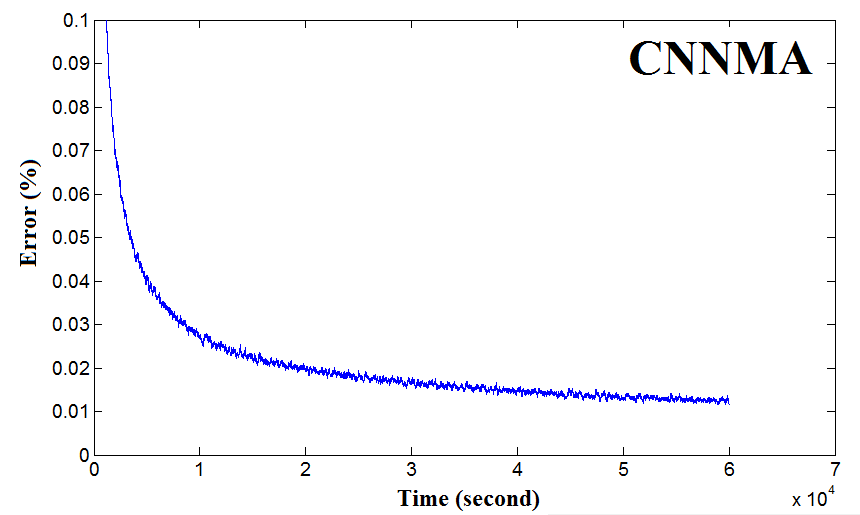}
\caption{Error vs Computation time for CNN by MA}
\label{fig:my_label}
\end{figure}

\begin{table}[t]
%\centering
\caption{Accuracy and Computation time for CNN, CNNMA}
\label{tab:my_label}

\begin{tabular}{c c c c c}
\hline
\multicolumn {1}{c}{\multirow{2}{*}{Epoch}} & \multicolumn{2}{c}{CNN} &
\multicolumn {2}{c}{CNNMA}\\
\cline {2-5}

\multicolumn {1}{r}{}       & \multicolumn {1}{c}{A1(\%)} & \multicolumn {1}{c}{T1(seconds)} &
\multicolumn {1}{c}{A2(\%)} & \multicolumn {1}{c}{T2(seconds)} \\
\hline

1		& 82.39 & 91.75	     & 86.99 $\pm$ 0.53   & 109.99 $\pm$ 10.47\\
2 		& 89.06 & 193.39 	 & 91.33 $\pm$ 0.43   & 203.04 $\pm$ 0.83  \\
3		& 91.13 & 297.31	 & 93.14 $\pm$ 0.31   & 302.84 $\pm$ 1.74  \\
4		& 92.33 & 379.44	 & 94.48 $\pm$ 0.15   & 402.42 $\pm$ 0.57  \\
5		& 93.11	& 479.04     & 95.11 $\pm$ 0.28   & 514.52 $\pm$ 7.54  \\
6		& 93.67 & 576.38	 & 95.72 $\pm$ 0.12   & 612.07 $\pm$ 2.33  \\
7 		& 94.25	& 676.57     & 95.99 $\pm$ 0.22   & 781.32 $\pm$ 23.57  \\
8		& 94.77 & 768.24	 & 96.26 $\pm$ 0.26   & 1062.11 $\pm$ 6.79 \\
9		& 95.37 & 855.85	 & 96.49 $\pm$ 0.27   & 1144.01 $\pm$ 79.84 \\
10		& 95.45 & 954.54	 & 96.89 $\pm$ 0.15   & 1274.00 $\pm$ 47.69 \\
\hline
100		& 98.65 & 10731	     & 98.75 & 17090 \\
\hline
\end{tabular}
\end{table}

In general, the experiments conducted for MNIST data set shown that the proposed methods are better than the original CNN, for any given epoch. As an example for the second epoch, the accuracy of original CNN is 89.06\%, while for CNNMA is 91.33\%. Accuracy improvement of the proposed method, compared to original CNN, varies of each epoch, with a range of values between 1.12\% (CNNMA, 9 epoch) up to 4.60\% (CNNMA, 1 epoch).

The computation time for the proposed method, compared to the original CNN, is in the range of \(1.02\times\) (CNNMA, three epochs : 302.84/297.31) up to \(1.38\times\) (CNNMA, eight epochs: 1062.11/768.24). 

%-------------------------------------------
\subsection{Experiment using CIFAR10 dataset}
%-------------------------------------------

The experiment of CIFAR10 dataset was conducted in MATLAB-R2014a, Ubuntu 14.04 LTS 64, on a PC with  Processor Intel Core i7-5820K, Four GPU GTX Titan X, Memory DDR2 RAM 64.00 GB, Hard disk 240 GB. The original program is MatConvNet from \cite{vedaldi}. In this paper, the program was modified with MA algorithm. The results can be seen in Fig. 8 for top-1 error and top-5 error.

The proposed method has proven very effective on CIFAR10 dataset with an accuracy of 99.6\%, for the last epoch in the top-1 error. In Table II different results from state of the art approaches are listed as a comparison. Another work proposed fine-tuning CNN using metaheuristic algorithm, harmony search (HS) \cite{Rosa} also compared in Table II. 

\begin{table}
\renewcommand{\arraystretch}{1.3}
\caption{State of the art CIFAR10 dataset}
\label{tab:example}
%\centering
\begin{tabular}{c|c}
    \hline
    Method  &  Accuracy (\%)\\
    \hline
    \hline

    \textbf{CNN-MA [ours]}    &   \textbf{99,14}\\
    \hline
    
    CNN-HS \cite{Rosa}    &   72.28\\
    \hline
    
    Fractional Pooling \cite{BG}   &   96.53\\
    \hline

    Large ALL-CNN  \cite{Jost}   &   95.59\\
    \hline
    
    Spatially Sparse CNN \cite{BG}   &   95,53\\
    \hline
    
    LSUV \cite{Dmytro}    &   94.16\\
    \hline
    
    CNN  \cite{matconvnet} &   80.46\\
    \hline

\end{tabular}
\end{table}

\begin{figure}
%\centering
\includegraphics[scale = 0.8]{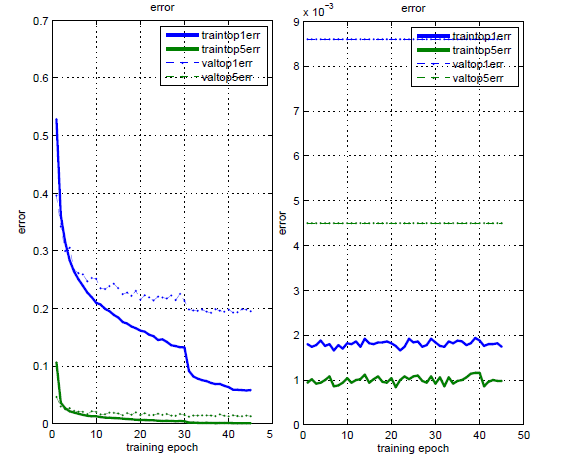}
\caption{Error for CNN and CNNMA }
\label{fig:my_label}
\end{figure}

%+++++++++++++++++++%
\section{Conclusion}
%+++++++++++++++++++%

This paper proposed a type of metaheuristic called Microcanonical Annealing algorithm to optimize the Convolutional Neural Network. Experimental result using MNIST and CIFAR-10 dataset demonstrated that although MA requires more computational time, the accuracy is reasonably better than the standard CNN without metaheuristic. This paper shows that on MNIST dataset, Microcanonical Annealing can improve the accuracy of Convolutional Neural Network, for all variations of epoch up to 4.60\%. The results obtained for CIFAR10 dataset, with an accuracy of 99.14\% (top-1 error), indicates that the proposed method is able to compete on the current state of the art approaches (96.53\%), in the field of image classification. For the future study, fining the proper MA parameters need to be investigated. Furthermore application of this proposed method using the other benchmark data set need to be explored, such as MMI and CKP facial expression data set, as well as ORI and ImageNet. For future research, we will investigate further on the computation time comparison between Microcanonical Annealing to the Simulated Annealing. We also need to examine further the accuracy on the CIFAR-10 dataset using other GPU-based deep learning frameworks such as Torch, Theano, Tensorflow, and Keras with more number of iterations.

%\nolinenumbers

%This is where your bibliography is generated. Make sure that your .bib file is actually called library.bib
\bibliography{library}

%This defines the bibliographies style. Search online for a list of available styles.
\bibliographystyle{abbrv}

\end{document}